\documentclass{article}

\usepackage{arxiv}

\usepackage[utf8]{inputenc} % allow utf-8 input
\usepackage[T1]{fontenc}    % use 8-bit T1 fonts
\usepackage{hyperref}       % hyperlinks
\usepackage{url}            % simple URL typesetting
\usepackage{booktabs}       % professional-quality tables
\usepackage{amsfonts}       % blackboard math symbols
\usepackage{nicefrac}       % compact symbols for 1/2, etc.
\usepackage{microtype}      % microtypography
\usepackage{lipsum}
\usepackage{graphicx}
\usepackage{float}

\usepackage{subcaption}
\usepackage{tikz}           % tikz package
\usepackage{pgfplots}
\usepackage{amsmath}
\usepackage{amssymb}
\usepackage{amsmath}
\RequirePackage{natbib}

\usepackage{graphics}
\usepackage{newunicodechar}
\newunicodechar{ﬁ}{fi}
\newunicodechar{ﬀ}{ff}

\def\O{\mathcal O}

\def\w{\mathbf w}
\def\H{\mathbf H}

\def\H{\mathbf H}
\def\x{\mathbf x}

\def\L{\mathcal{L}}

\def\sumi{\sum_{i=1}^n}

\def\t{^\top}

\def \L {\mathcal{L}}

\title{Importance of Data Loading Pipeline in Training Deep Neural Networks}

\author{
Mahdi Zolnouri \& Xinlin Li \& Vahid Partovi Nia  \\
Huawei Noah's Ark Lab \\
Montreal, QC H3N 1X9, Canada \\
\texttt{\{mahdi.zolnouri,xinlin.li1,vahid.partovinia\}@huawei.com} \\
}

\begin{document}
\maketitle

\begin{abstract}
Training large-scale deep neural networks is a long, time-consuming operation, often requiring many GPUs to accelerate. In large models, the time spent loading data takes  a significant portion of model training time. As GPU servers are typically expensive, tricks  that can save training time are valuable. Slow training is observed especially  on real-world applications where exhaustive data augmentation operations are required. Data augmentation techniques include: padding, rotation, adding noise, down sampling, up sampling, etc. These additional operations increase the need to build an efficient data loading pipeline, and to explore existing tools to speed up training time. We focus on the comparison of two main tools designed for this task, namely binary data format to accelerate \emph{data reading}, and  NVIDIA DALI to accelerate \emph{data augmentation}. Our study shows improvement on the order of $20\%$ to $40\%$ if such dedicated tools are used.
\end{abstract}

% keywords can be removed
% \keywords{First keyword \and Second keyword \and More}

\section{Introduction}

Deep neural networks have achieved great successes in various domains such as computer vision \cite{chollet2017xception, badrinarayanan2017segnet}, natural language processing \cite{goldberg2016primer, devlin2018bert}, and speech recognition \cite{povey2011kaldi} among others. This is a result of deeper and wider models,  which allow modeling large and complex data. As computing hardware has improved, larger data sets are analyzed. It appears that processing power always falls behind data volume and model size. In order to make the training process more efficient, several fields  are developing new techniques such as providing dedicated tools to accelerate training and inference, as well as neural model compression to deploy a simpler model with comparable accuracy but fewer operations. 

Training acceleration is a difficult task. Let's understand the core of the problem using an a very simple neural network, e.g. logistic regression. Suppose $N$ pairs of input features of dimension $d$, say $\x_i$ and binary output data, say $y_i$ are available $(\x_i, y_i), i=1,\ldots, N$. A logistic regression model is equivalent to a fully-connected network with a single hidden layer and a single neuron. As the data size gets bigger in terms of $N$ and $d$, training requires more computation.

A training process optimizes a loss function, here

\begin{equation}
\begin{aligned}
    \L(\w) =& - \sumi y_i \log \sigma(\x_i\t \w) \\
    &+ (1-y_i) \log (1-\sigma(\x_i\t\w)),
\end{aligned}
\end{equation}

where $$\sigma(x)=\{1+\exp(-x)\}^{-1}$$ is the sigmoid activation. For the case of logistic regression iterative re-weighted least squares is often used to  optimize $\L$, which is equivalent to Newton's method. Newton's method starts from an initial estimate $\w_0$ and updates 
\begin{equation}
\w_{t+1} \leftarrow \w_t  - \H^{-1} \nabla \L(\w)\mid_{\w=\w_t}
\label{eq:newton}
\end{equation}
where $\H$ is the $d\times d$ Hessian of $\L$ and $\nabla \L = \sumi \nabla \L_i(\w)$ is the sum of individual gradients, each of length $d$.  Computing the local approximation of $\H = \sumi \x_i\x_i\t  $  is  of $\O(nd^2)$ and factorizing it is of $\O(nd^3)$, if not impossible.

Increasing the number of data points  $N\to\infty$ theoretically makes the optimization easier, because $\L(\w)$ has more curvature as $N$ increases, a \emph{blessing}. However, computation of $\L(\w)$, $\nabla \L(\w)$, and $\H$ becomes a \emph{curse}  since all of these quantities are in the form of a sum and their computation  may lead to memory overflow. Optimization using Newton's method becomes increasingly hard with large feature size $d$. Large features are common in almost all machine learning challenges. The remedy for large $N$ is to break computations into smaller sub-operations. The remedy for large $d$ is to switch from a second-order approximation, i.e., Newton's method, to a first-order approximation, i.e., the gradient method. 

Computing partial sums is a simple way to overcome the large $N$ issue, so that each partial sum remains within the memory resource constraints. Then the final quantity is computed by summing the partial sums, perhaps with a proper re-scaling. The idea of partial sum is somehow a re-shape of the mini batch training approach.

In neural networks with a large $d$ and $n$, numerical optimization is simplified to gradient descent  in which the hessian $\H$  is replaced by the identity matrix with a positive scalar learning rate  $$\H= {1 \over \eta} \mathbf I, \eta>0$$ 
so the weight update is simplified to
\begin{equation}
\w_{t+1} \leftarrow \w_t  - \eta \nabla \L(\w)\mid_{\w=\w_t}.
\end{equation}

Furthermore, to benefit from parallel computation, $N$ data points are arranged in $n$ random mini batches, each of size $B$, i.e. $N=nB.$ Each batch has its own gradient  
$$\nabla \L_b(\w) = \sum_{i=1}^B \nabla \L_{bi}(\w),$$ 
which is equivalent to scaling $\eta$ by $n$, on average. This allows computations to be run in parallel for each batch \cite{cotter2011better, muralidhar2006data}.

Even if computations are run in parallel, all data still needs to be  fed to the optimizer in  several rounds of \emph{epochs}, similar to Newton's method.  Therefore, investing in an efficient data loading pipeline plays an important role in training speed \cite{yang2019accelerating}. 

The rest of the paper focuses on clarifying the benefit of a dedicated tool such as DALI\footnote{\url{https://github.com/NVIDIA/DALI}}  for managing data loading implemented by NVIDIA in PyTorch, while a  using a convenient data reading format such as Hierarchical Data Format 5  (HDF5) \cite{koranne2011hierarchical} or TensorFlow Record \cite{abadi2016tensorflow} to accelerate file reading.

\begin{figure}[H]
\centering
\includegraphics[width=1.0\textwidth]{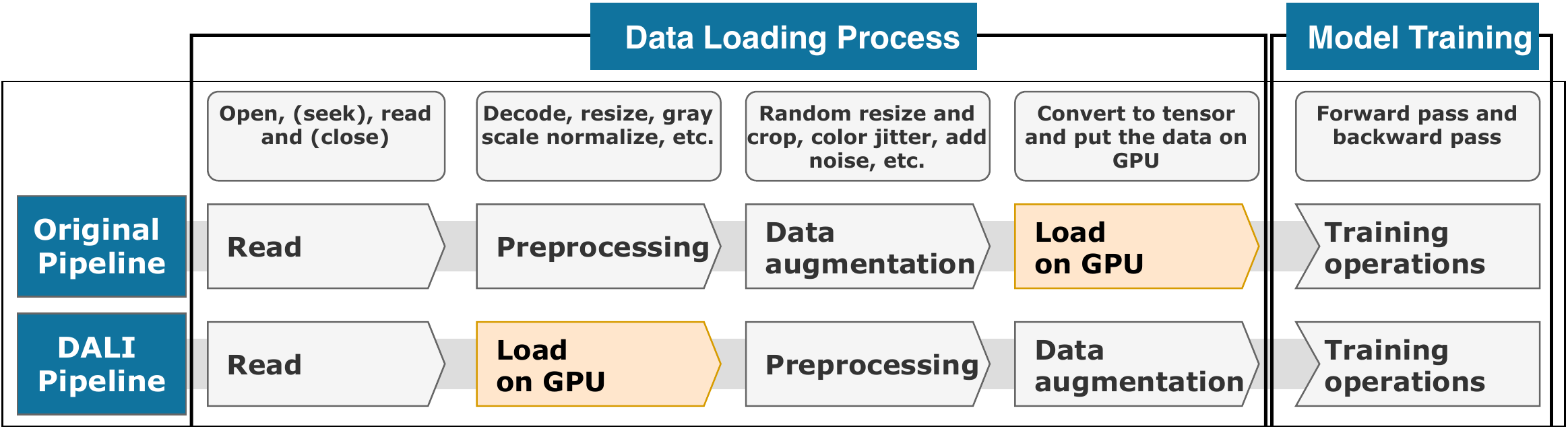}
\caption{Batch training sequence performed on a GPU with and without DALI.}
\label{fig:batchtraining}
\end{figure}

Data loading is a crucial part of model training in neural networks. It begins by reading the data from a secondary memory storage, such as \emph{Solid State Drive}, then caches it into a primary memory storage, such as \emph{Random Access Memory}. This data transfer includes extra  operations like data augmentation to feed the  data to the model. See Figure~\ref{fig:batchtraining}. 

There are two main issues in the data loading process i) reading data directly from files is inefficient ii) resource allocation for extra operations on data overloads the CPU.

Reading data from individual files directly is slow. This happens when the entire data set is not cached in the local memory. Every single file is opened, is read, and is closed sequentially. These sequential operations add considerable  overhead to the file retrieval time. One solution to prevent this overhead is to use the Hierarchical Data Format version 5 (HDF5) \cite{koranne2011hierarchical}, which has an open-source library to store, manipulate, and manage the large data set. HDF5 format stores multiple data sets in one ﬁle as a multidimensional array of binary data. It also groups storage layout by  storing data in ﬁxed-size chunks on disk. Another alternative to HDF5 is the TensorFlow Record (TFRecord) \cite{abadi2016tensorflow}, which uses a sequence of binary strings to store data. It allows large data sets to be sequentially loaded to the local memory. 

\begin{figure}[H]
\centering
\includegraphics[width=0.5\textwidth]{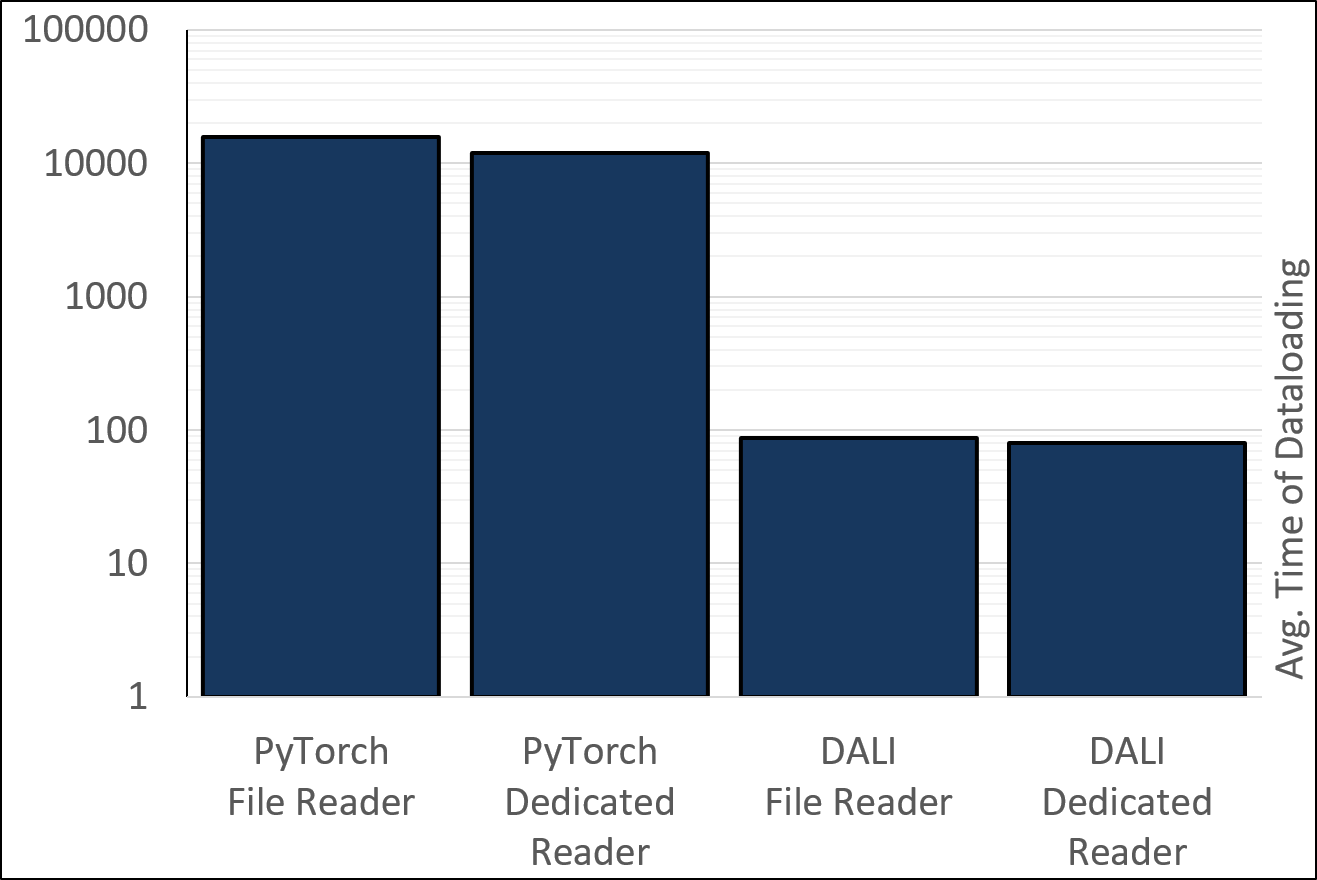}
\caption{Logarithmic scale of average time of data loading (in micro second) for ResNet-50 \cite{he2016deep} on ImageNet data set with batch-size  $B=256$}
\label{fig:timecompare}
\end{figure}

The whole process of data loading is managed by CPU, which can create a bottleneck for model training. This bottleneck happens normally in the case of multi-node-multi-GPU training, as loading batches of data takes more  time than forward-backward propagation. To prevent the CPU bottleneck issue, NVIDIA Data Loading Library (DALI) helps by sharing some data loading tasks between the CPU and GPU, to prevent the CPU bottleneck issue. Figure~\ref{fig:timecompare} shows how employing this library  improves the data loading process considerably. DALI is a collection of highly optimized building blocks and execution engines which provides a full data accelerated pipeline: from reading the data to preparing for  training and inference. 

\section{Data Format}

In order to explore the effect that data format and resource allocation has on data loading performance, we present four pipelines for data loading: with/without dedicated reader, and with/without resource allocation. We used PyTorch  version 1.2.0  to implement these pipelines on a computer vision task.

The PyTorch \emph{file reader} pipeline is the PyTorch \emph{DataLoader} class which combines \emph{data set} object with a  \emph{sampler} object, to provide a single or multi-process iterators over the data set. This pipeline reads data from individual JPEG files on the storage and uses the PyTorch \emph{Transforms} class to chain several image transformation operations together.  This prepares the data for training by doing the data augmentation. By default, all these operations are directed to the  CPU. 

The \emph{dedicated reader}  pipeline is also based on the PyTorch \emph{DATALOADER} class. This pipeline stores data differently, i.e. instead of reading data from individual JPEG files, the entire train and validation data sets are stored as two HDF5 files.

\section{Resource Allocation}
This pipeline uses the NVIDIA DALI library to read data from JPEG files, process and then feed GPUs for training. In the DALI pipeline, the data loading process is shared between CPU and GPU. This means that all operations on data, such as resizing, cropping and data augmentation can be run on CPU, GPU, or a mix of both. To measure the effect of file formatting, data file retrieval has been done in two cases: reading directly from JPEG files and using a dedicated file reader to read data from TFRecord data set. 

\section{Training Time Improvement}
We summarize our experiments using  four configurations, i.e., with/without a dedicated file reader, and with/without DALI. In order to compare these pipelines, we performed several experiments in two use cases:  with few or with extensive \emph{data augmentation} operations. By few operations, we mean resize with random crop and random horizontal flip operations and by extensive, we mean resize with random crop operation, random horizontal flip operation and random adjustment of the brightness, contrast and saturation of an image. Simple data augmentation  is applied in most deep learning prototypes, while extensive data augmentation is very common in  industry to ensure model robustness in real products.  Our experiments are run twice, once  on small subset of InsightFace \cite{deng2019retinaface, guo2018stacked, deng2018menpo, deng2018arcface} and once on large ImageNet data set \cite{deng2009imagenet}.

\begin{figure}
\centering 
\includegraphics[width=0.35\textwidth]{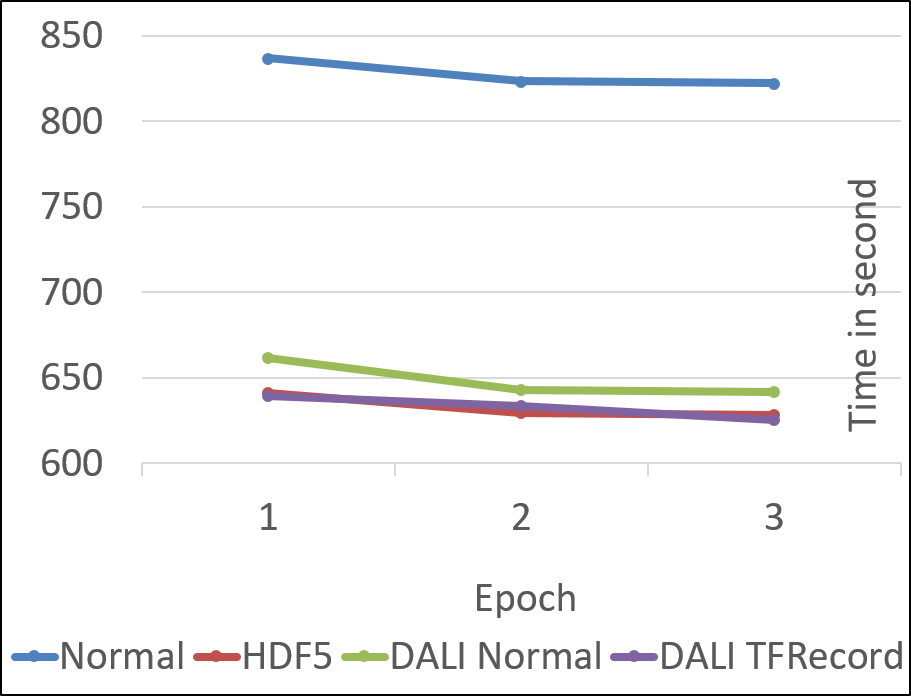}\includegraphics[width=0.35\textwidth]{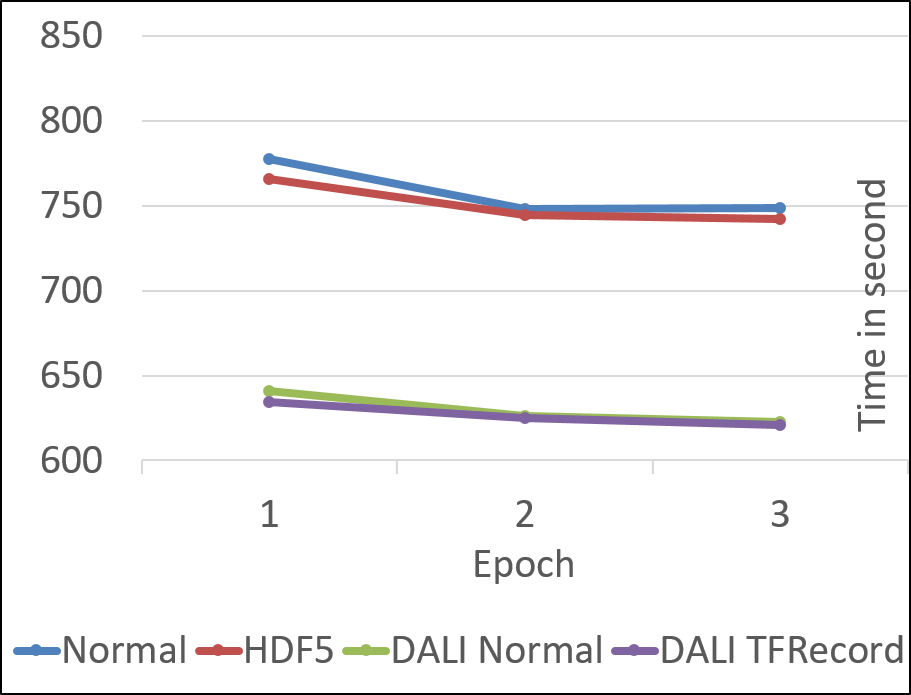}\\
\includegraphics[width=0.35\textwidth]{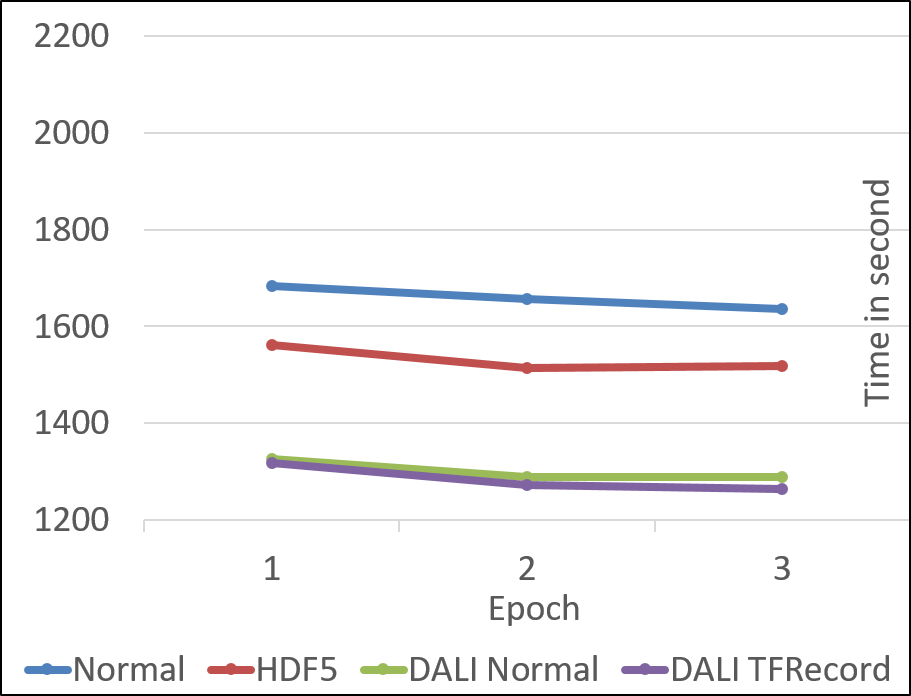}\includegraphics[width=0.35\textwidth]{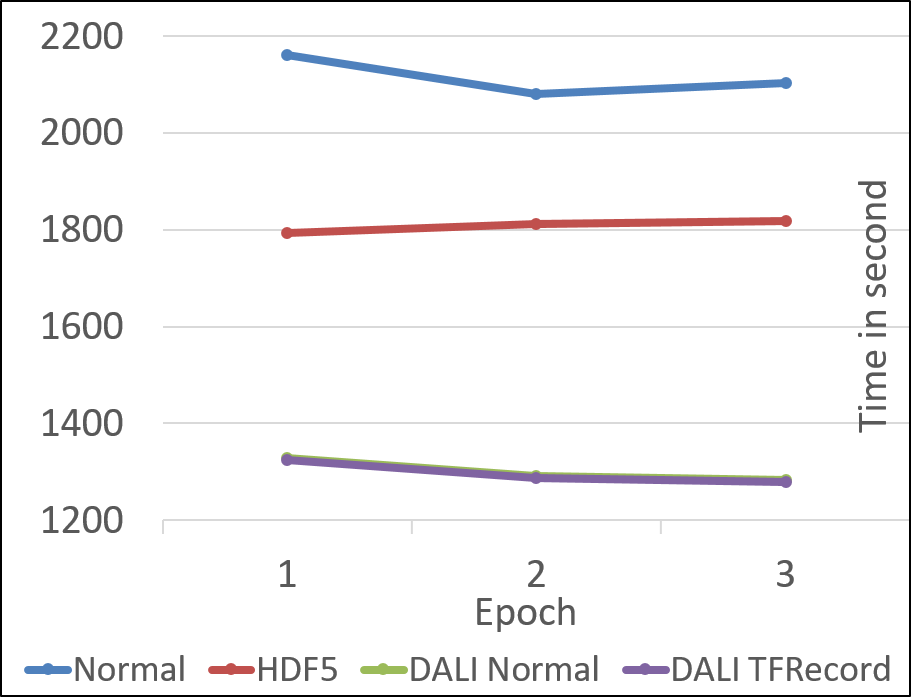}
\caption{Epoch time for the small subset of InsightFace (top panels) and large ImageNet data (bottom panels), while few data processing operations are performed (left panels) and while extensive preprocessing operations are performed (right panel). The experiments in top panels were performed on 4 GPUs (NVIDIA TITAN 12 GB memory) and in bottom panels on 8 GPUs (NVIDIA TESLA V100 32GB memory). } 
\label{fig:lineplot}
\end{figure}

Figure~\ref{fig:lineplot} shows that a dedicated data reader is enough to improve epoch time, but only  if data is small. However, the DALI pipeline still is a winner for small data requiring extensive data augmentation.  This is because DALI distributes the data augmentation operation between CPU and GPU to avoid  CPU overcharge. This can be seen by comparing the top left panel, with the top right panel.

\begin{figure}
\begin{minipage}[t]{0.4\linewidth}
\includegraphics[width=\linewidth]{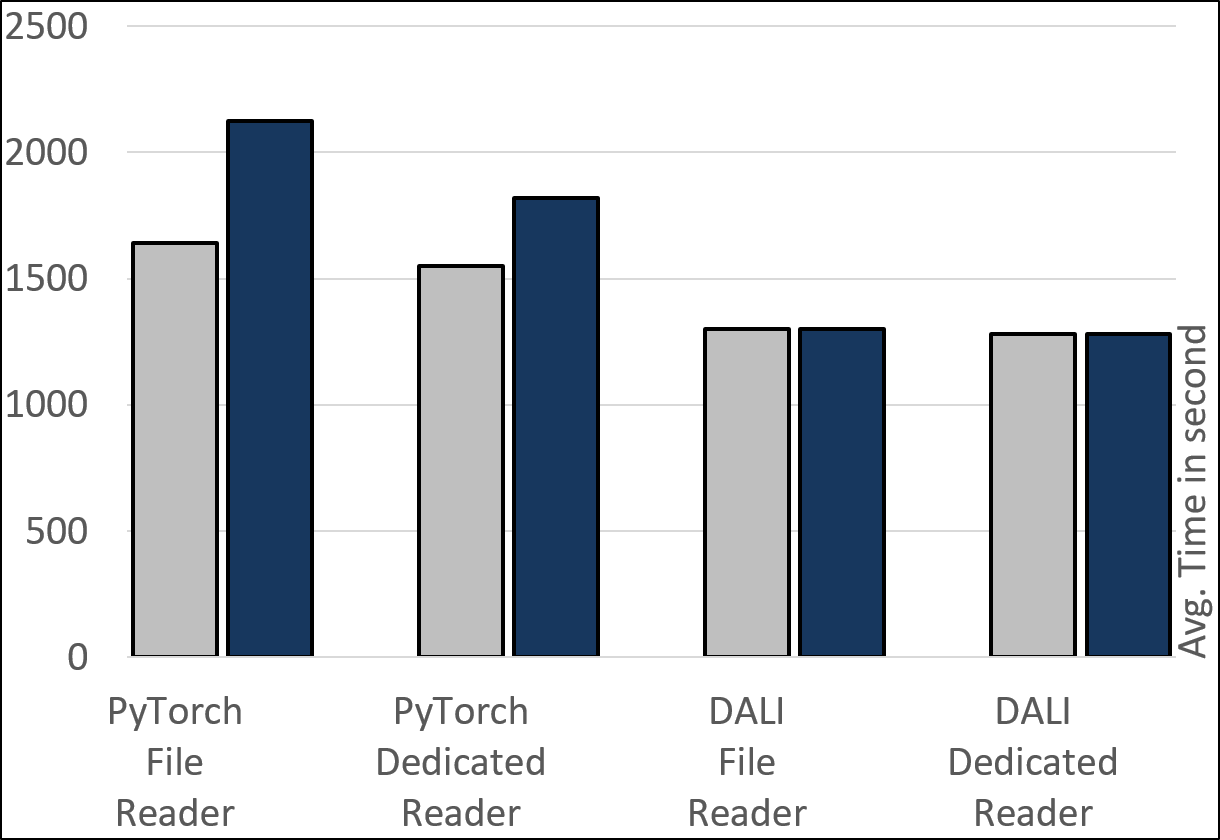}
\caption{Time comparison for few (gray) and extensive (dark) data augmentations.}
\label{fig:timecompare2}
\end{minipage}
\hfill
\begin{minipage}[t]{0.4\linewidth}
\includegraphics[width=\linewidth]{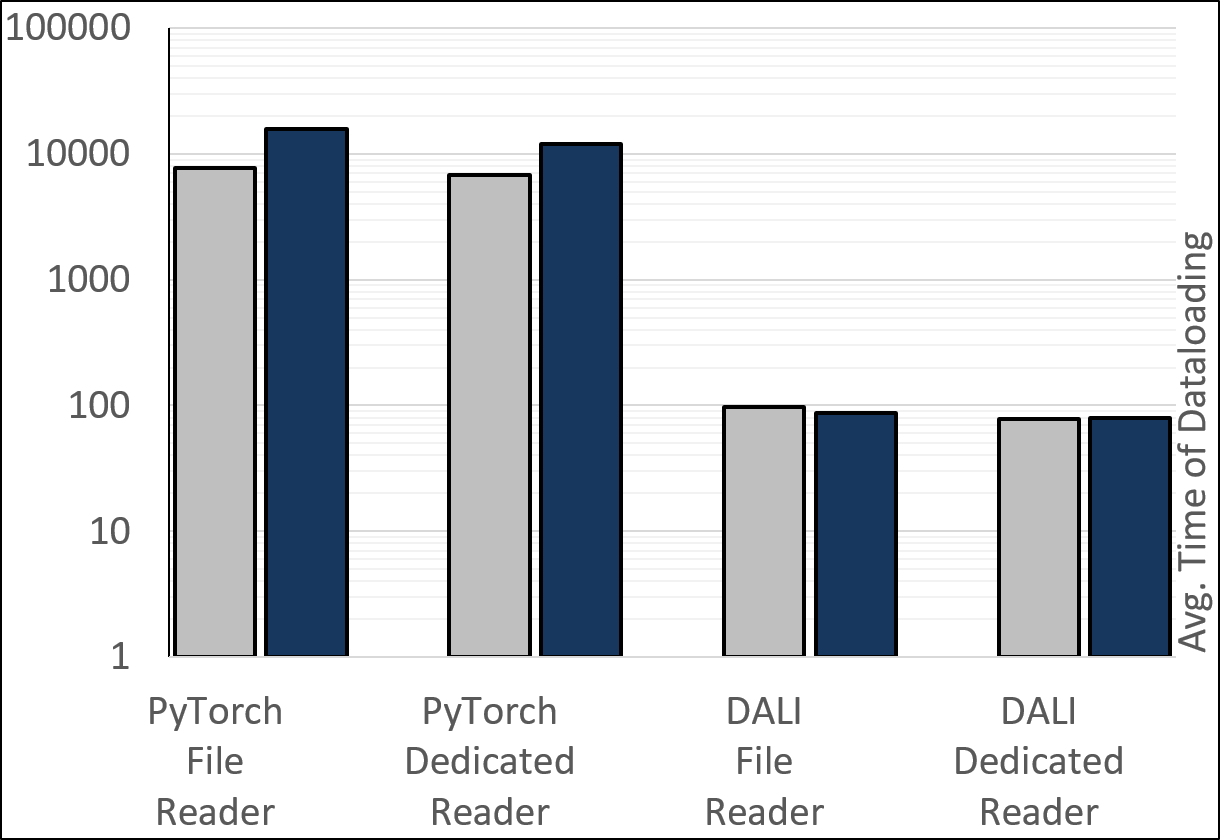}
\caption{logarithmic scale plot of average data loading time (per epoch), four data loader pipelines: with/without DALI, and with/without a dedicated reader. Time for few (gray) and extensive (dark) data augmentations.}
\label{fig:logscale}
\end{minipage}
\end{figure}

\begin{figure}
\begin{minipage}[t]{0.4\linewidth}
\includegraphics[width=\linewidth]{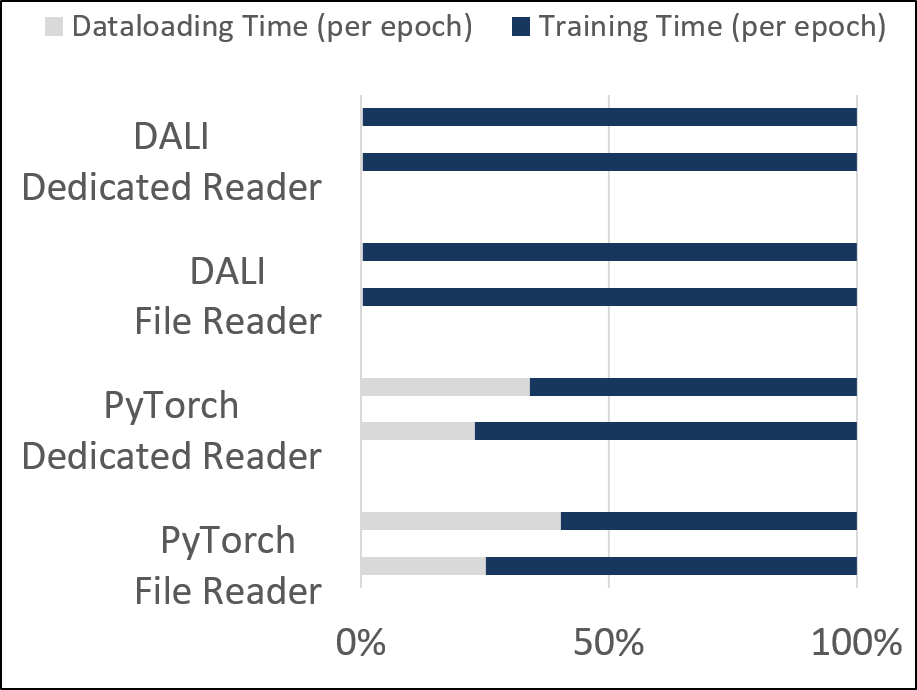}
\caption{Percentage of data loading time (gray) added on top of training time (dark) in a single training epoch. Each data loader test case has two bar charts for showing few and extensive data augmentations.}
\label{fig:percentplot}
\end{minipage}%
\hfill
\begin{minipage}[t]{0.4\linewidth}
\includegraphics[width=\linewidth]{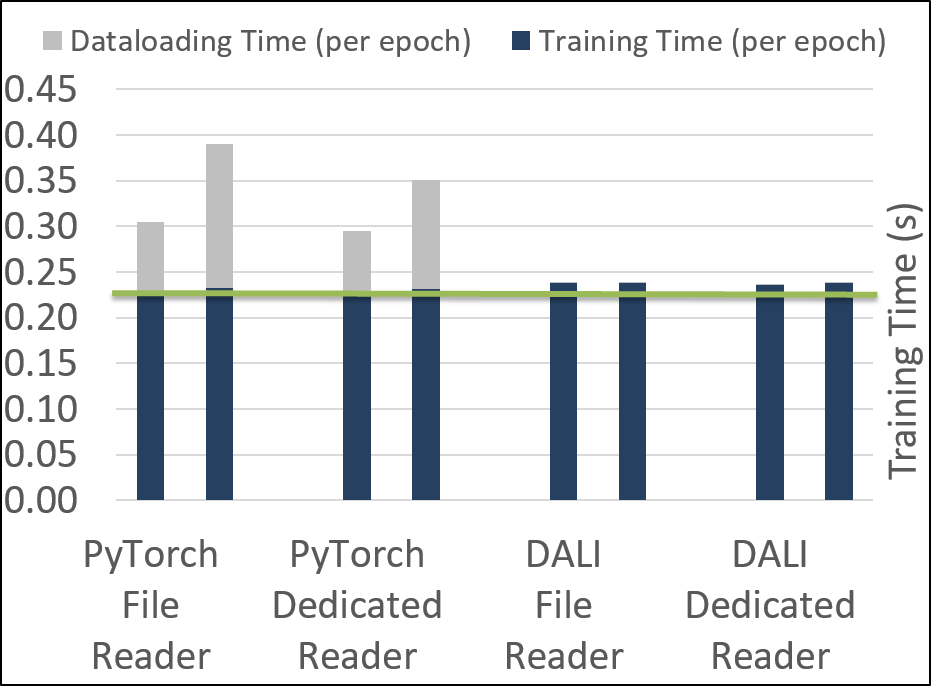}
\caption{Training time on four data loader pipelines: with/without DALI, and with/without a dedicated reader. Each data loader pipeline has two bar charts for showing few and extensive data augmentations.}
\label{fig:stackedbar}
\end{minipage}%
\end{figure}

If  extensive data preprocessing operations are performed in a larger ImageNet data, CPU workload becomes the bottleneck even with few data augmentation operations. Consequently, epoch time increases in data loading  and DALI can avoid the CPU overcharge by performing some of these operations on GPU. Figure~\ref{fig:lineplot} (bottom right) shows that DALI improves data loading from $2200$ seconds to $1300$ seconds, to  gain performance benefit of about $40\%$. Using only a dedicated reader without DALI improves the training time from $1800$ seconds to $1300$ seconds, giving a performance improvement of $30\%$. This effect is also visible in Figure~\ref{fig:timecompare2} on a single epoch time. DALI fuses  multiple operations such as cropping and normalizing and run it on one GPU CUDA kernel. This speeds up data augmentation process by reducing the number of memory access.

\section{Data Loading Improvement}
Let's move from epoch time to data loading time by removing forward and backward pass from training time, see Figure~\ref{fig:batchtraining}. Data loading contributes about $40\%$ to the epoch time, see Figure~\ref{fig:percentplot}. 

Figure~\ref{fig:logscale} confirms that the DALI pipeline considerably improves data loading by a factor of $100 \times$. However, for very large models, the GPU is only required to perform forward and backward passes, so loading the CPU for data is wiser. Therefore, it is important to keep the CPU and GPU load well balanced through DALI load option.  Figure~\ref{fig:stackedbar} confirms the same message when training time is stacked on data loading time to measure the epoch time overall.

In Figure~\ref{fig:stackedbar} data loading is stacked on training to reflect the epoch time overall. There is a small difference between the common data loader pipeline and NVIDIA DALI data loader pipeline. Using the GPU resources for data loading may slow down the overall model training time if it fails  to balance CPU and GPU load. 

By default, DALI uses the first GPU slot to perform data loading process. However, the NVIDIA APEX library uses multiple GPUs for this task. This flexibility becomes increasingly important for large models while all data and models cannot be loaded into a single GPU and multi-GPU operation becomes a necessity. 
\section{Conclusion}
NVIDIA DALI provides an effective alternative to common  data loading process. It provides a full pipeline of optimizations including data readers and tools to accelerate training and inference. It also enables most data augmentation  operations to be performed on GPU and on CPU. In addition, DALI prepared a full pipeline for common data sets like MS-COCO data set \cite{lin2014microsoft} as well as provides a reader for TFRecord and CAFFE LMDB data formats  \cite{jia2014caffe}. DALI remains extremely flexible by supporting ExternalSource operator so that implementation of unsupported readers such as HDF5 becomes feasible.

Here, we focused on large models, in which DALI GPU improves training time. However, training is faster with DALI CPU for small networks.

% conference papers do not normally have an appendix

% use section* for acknowledgement
\section*{Acknowledgment}
The authors would like to thank Eyy\"ub Sari and Vanessa Courville for their assistance in this project. We appreciate fruitful technical discussions with Huawei Cloud Core Shanghai   Gang  Chi  and  Pengcheng  Tang as well as Jiajin Zhang from Noah's Ark Shenzhen  engineering team. 

\bibliographystyle{iclr2020}  
\bibliography{references}  

\end{document}